\documentclass[10pt,twocolumn,letterpaper]{article}

\usepackage{iccv}
\usepackage{times}
\usepackage{epsfig}
\usepackage{graphicx}
\usepackage{amsmath}
\usepackage{amssymb}
\usepackage{booktabs}
\usepackage{paralist}
\usepackage{multirow}
\usepackage{enumitem}
\usepackage{subcaption}
\usepackage{wrapfig}
\usepackage{mysymbols}

\newcommand{\myquote}[1]{\emph{`#1'}}
\newcommand{\myapprox}{{\raise.17ex\hbox{$\scriptstyle\sim$}}}
\newcommand{\xhdr}[1]{\vspace{2pt}\noindent\textbf{#1}}

\usepackage[toc,page]{appendix}
\usepackage[usenames,dvipsnames]{xcolor}

\usepackage[pagebackref=true,breaklinks=true,letterpaper=true,colorlinks,bookmarks=false]{hyperref}

\iccvfinalcopy %

\def\iccvPaperID{706} %

\ificcvfinal\fi

\belowdisplayskip \abovedisplayskip

\newlength{\abstractReduceTop}
\newlength{\abstractReduceBot}

\newcommand{\reducedSection}[1]{\vspace{\sectionReduceTop}\section{#1}\vspace{\sectionReduceBot}}
\newlength{\sectionReduceTop}
\newlength{\sectionReduceBot}

\newcommand{\reducedSubSection}[1]{\vspace{\subsectionReduceTop}\subsection{#1}\vspace{\subsectionReduceBot}}
\newlength{\subsectionReduceTop}
\newlength{\subsectionReduceBot}

\newlength{\subsubsectionReduceTop}
\newlength{\subsubsectionReduceBot}

\newlength{\captionReduceTop}
\newlength{\captionReduceBot}

\newlength{\eqnReduceTop}
\newlength{\eqnReduceBot}

\newlength{\horSkip}
\newlength{\verSkip}

\newlength{\figureHeight}
\begin{document}

\title{Taking a HINT: Leveraging Explanations to Make \\ Vision and Language Models More Grounded}

\author{
	Ramprasaath R. Selvaraju$^1$ \hspace{1pc}
    Stefan Lee$^{1,4}$ \hspace{1pc}
    Yilin Shen$^2$ \hspace{1pc}
	Hongxia Jin$^2$ \\
	Shalini Ghosh$^2$ \hspace{1pc}
	Larry Heck$^2$ \hspace{1pc}
    Dhruv Batra$^{1,3}$ \hspace{1pc}
    Devi Parikh$^{1,3}$ \\
    \hspace{-1pc}$^1$Georgia Institute of Technology,
	$^2$Samsung Research,
	$^3$Facebook AI Research,
	$^4$Oregon State University\\
	{\tt\small \{ramprs, steflee, dbatra, parikh\}@gatech.edu}\\
    {\tt\small \{yilin.shen, hongxia.jin, shalini.ghosh, larry.h\}@samsung.com}\\
}

\maketitle

\thispagestyle{plain}
\pagestyle{plain}

\begin{abstract}
Many vision and language models suffer from poor visual grounding -- often falling back on easy-to-learn language priors rather than basing their decisions on visual concepts in the image. In this work, we propose a generic approach called Human Importance-aware Network Tuning (HINT) that effectively leverages human demonstrations to improve visual grounding. HINT encourages deep networks to be sensitive to the same input regions as humans. Our approach optimizes the alignment between human attention maps and gradient-based network importances -- ensuring that models learn not just to look at but rather rely on visual concepts that humans found relevant for a task when making predictions. 
We apply HINT to Visual Question Answering and Image Captioning tasks, outperforming top approaches on splits that penalize over-reliance on language priors (VQA-CP and robust captioning) using human attention demonstrations for just 6\% of the training data. 
\end{abstract}

\reducedSection{Introduction}

Many popular and well-performing models for multi-modal, vision-and-language tasks exhibit poor visual grounding -- failing to appropriately associate words or phrases with the image regions they denote and relying instead on superficial linguistic correlations \cite{agrawal2018don, vqa-ba,YinYang,goyal2016making,johnson2016clevr}. For example, answering the question \myquote{What color are the bananas?} with yellow regardless of their ripeness evident in the image. When challenged with datasets that penalize reliance on these sort of biases \cite{agrawal2018don,goyal2016making}, state-of-the-art models demonstrate significant drops in performance despite there being no change to the set of visual and linguistic concepts about which models must reason.  

In addition to these diagnostic datasets, another powerful class of tools for observing this shortcoming has been gradient-based explanation techniques \cite{sundararajan2017axiomatic,zhang2016top,springenberg2014striving,gradcam_ijcv} which allow researchers to examine which portions of the input models rely on when making decisions. Application of these techniques has shown that vision-and-language models often focus on seemingly irrelevant %
image regions that differ significantly from where human subjects fixate when asked to perform the same tasks \cite{vqahat,gradcam_ijcv} -- \eg focusing on a produce stand rather than the bananas in our example.

\begin{figure}[t]\label{fig:hint_teaser}
\centering
\includegraphics[width=0.95\columnwidth]{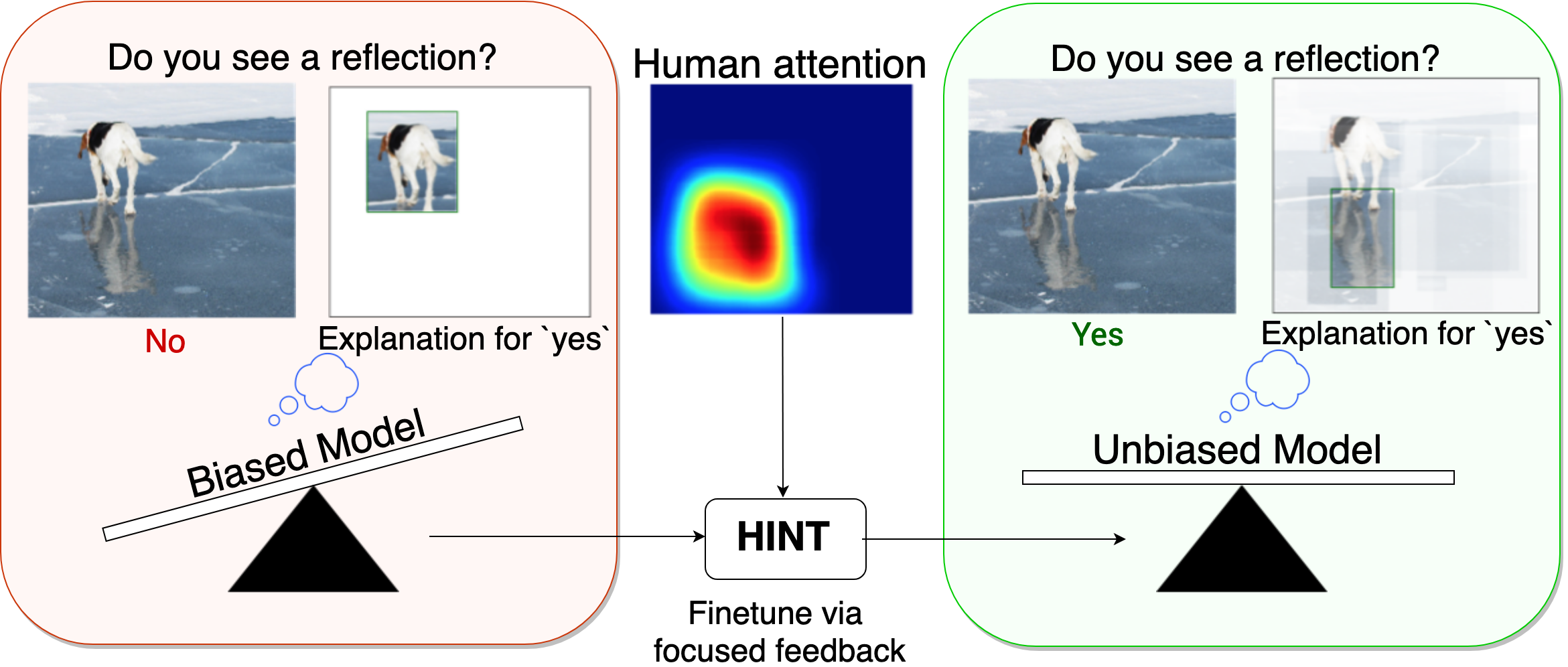}
\caption{Our approach, HINT, aligns visual explanations for output decisions of a pretrained model with spatial input regions deemed important by human annotators -- forcing models to base their decisions on these same region and reducing model bias.}
\vspace{-10pt}
\end{figure}

While somewhat dissatisfying, these findings are not entirely surprising -- after all, standard training protocols do not provide any guidance for visual grounding. Instead, models are trained on input-output pairs and must resolve grounding from co-occurrences  -- a challenging task, especially in the presence of more direct and easier to learn correlations in language. Consider our previous example question, the words `color', `banana', and `yellow' are given as discrete tokens that will trivially match in every occurrence when these underlying concepts are referenced. In contrast, actually grounding this question requires dealing with all visual variations of bananas and learning the common feature of things described as `yellow'. 
To address this, we explore if giving a \emph{small hint} in the form of human attention demonstrations can help improve grounding and reliability. %

For the dominant paradigm of vision-and-language models that compute an explicit question-guided attention over image regions \cite{shih_2016,xiong2016dynamic,kazemi_2017,yang2016stacked,lu2018Neural,anderson_2017}, a seemingly straight-forward solution is to provide explicit grounding supervision -- training models to attend to the appropriate image regions. While prior work \cite{qiao2018exploring,liu2017attention} has shown this approach results in more human-like attention maps, our experiments show it to be ineffective at reducing language bias. Crucially, attention mechanisms are bottom-up processes that feed final classification models such that \emph{even when attending to appropriate regions, models can ignore visual content in favor of language bias.}
In response, we introduce a generic, second-order approach that instead aligns gradient-based explanations with human attention.

Our approach, which we call Human Importance-aware Network Tuning (HINT), enforces a ranking loss between human annotations of input importance and gradient-based explanations produced by a deep network -- updating model parameters via a gradient-of-gradient step. Importantly, this constrains models to not only look at the correct regions but to also be sensitive to the content present there when making predictions. 
While we experiment with HINT in the context of vision-and-language problems, the approach itself is general and can be applied to focus model decisions on specific inputs in any context.

We apply HINT to two tasks -- Visual Question Answering (VQA) \cite{antol2015vqa} and image captioning \cite{LinECCV14coco} -- and find our approach significantly improves visual grounding. 
With human importance supervision for only 6\% of the training set, our HINT'ed model improves the state-of-the-art by 8 percentage points on the challenging dataset VQA Under Changing Priors (VQA-CP)~\cite{agrawal2018don}, which is designed to test visual grounding. 
In both VQA and Image Captioning, we see significantly improved correlations between human attention and visual explanations for HINT trained models, showing that models learn to make decisions using similar evidence as humans (even on new images). 
We perform human studies which show that humans perceive models trained using HINT to be more reasonable and trustworthy. %

\xhdr{Contributions.} To summarize our contributions, we
\begin{compactitem}[\hspace{2pt}\textbullet]
\item introduce Human Importance-aware Network Tuning (HINT), a general approach for constraining the sensitivity of deep networks to specific input regions and demonstrate it results in significantly improved visual grounding for two vision and language tasks, 
\item set a new state-of-the-art on the bias-sensitive VQA Under Changing Priors (VQA-CP) dataset \cite{agrawal2018don}, and 
\item conduct studies showing that humans find HINTed models more trustworthy than standard models. 
\end{compactitem}

\reducedSection{Related Work}
\label{sec:rel}

\noindent 
\textbf{Model Interpretability.}~ 
There has been significant recent interest in building machine learning models that are 
transparent and interpretable in their decision making process.
For deep networks, several works propose explanations 
based on internal states of the network \cite{zeiler_eccv14,GoyalMPB16,Zhou2014ObjectDE,gradcam_ijcv}. 
Most related to our work is the approach of Selvaraju \etal\cite{gradcam_ijcv} which computes neuron importance as part of a visual explanation. 
In this work, we enforce that these importance scores align with importances provided by domain experts.

\xhdr{Vision and Language Tasks.}
Image Captioning \cite{Lin_2014} and Visual Question Answering (VQA) \cite{antol2015vqa} have emerged as two of the most widely studied vision-and-language problems. The image captioning task requires generating natural language descriptions of image contents and the VQA task requires answering free-from questions about images. 
In both, models must learn to associate image content with natural free-form text. Consequentially, attention based models that explicitly reason about image-text correspondences have become the dominant paradigm \cite{shih_2016,xiong2016dynamic,kazemi_2017,yang2016stacked,lu2018Neural,anderson_2017}; however, there is growing evidence that even these attentional models still latch onto language biases \cite{agrawal2018don, YinYang, anne2018women}.

Recently, Agrawal \etal \cite{agrawal2018don} introduced a novel, bias-sensitive dataset split for the VQA task. This split, called VQA Under Changing Priors (VQA-CP), is constructed such that the answer distributions differ significantly between training and test. As such, models that memorize language associations in training instead of actually grounding their answers in image content will perform poorly on the test set. Likewise Lu \etal \cite{lu2018Neural} introduce a robust captioning split of the COCO captioning dataset \cite{Lin_2014} in which the distribution of co-occurring objects differs significantly between training and test. We use these dataset splits to evaluate the impact of our method on visual grounding.

\xhdr{Debiasing Vision and Language Models.}
A number of recent works have aimed to reduce the effect of language bias in vision and language models. 

Hendricks \etal \cite{anne2018women} study the generation of gender-specific words in image captioning -- showing that models nearly always associated male gendered words to people performing extreme sports like snowboarding regardless of the image content. Their presented Equalizer approach encourages models to adjust their confidence depending on the evidence present -- confident when gender evidence is visible and unsure when it is occluded by ground-truth segmentation masks. Experiments on a set of captions containing people show this approach reduces gender bias.

For VQA, Agrawal \etal \cite{agrawal2018don} developed a Grounded VQA model (GVQA) that disentangles the vision and language components -- consisting of separate visual concept and answer cluster classifiers. 
This approach uses a question's type (\eg ``What color ...'') to determine the space of possible answers and the question target (\eg ``banana'') to detect visual attributes in the scene that are then filtered by the possible answer set. 
While effective, this requires multi-stage training and is difficult to extend to new models. 
Ramakrishnan \etal \cite{advregvqa_nips_2018} introduce an adversarial model agnostic regularization technique to reduce bias in VQA models -- pitting the model against a question-only adversary.

\xhdr{Human Attention for VQA.} 
Das \etal \cite{vqahat} collected human attention maps for a subset of the VQA dataset~\cite{antol2015vqa}. Given a question and a blurry image, humans were asked to interactively deblur regions in the image until they could confidently answer.  In this work, we utilize these maps, enforcing the gradient-based visual explanations of model decisions to closely match the human attention.

\xhdr{Supervising model attention.} 
Liu \etal \cite{liu2017attention} and Qiao \etal \cite{qiao2018exploring} apply human attention supervision to attention maps produced by the model for image captioning and VQA, respectively. 
We experiment with a similar approach but find that the improved attention correlation does not translate to reduced reliance on language bias -- 
even with appropriate model attention, the remaining network layers can still disregard the visual signal in the presence of strong biases. 
We also show how gradient explanations are more faithful to model decisions by directly linking model decisions input regions, so that aligning these importances ensures the model is basing its decision on human-attended regions. 

\xhdr{Aligning gradient-based importances.}
 Selvaraju \etal \cite{niwt} proposed an approach to learn a mapping between gradient-based importances of individual neurons within a deep network (from \cite{gradcam_ijcv}) and class-specific domain knowledge from humans in order to learn classifiers for novel classes. 
In contrast, we align gradient-based importances to human attention maps to improve network grounding.

\reducedSection{Preliminaries}

While our approach is general-purpose and model agnostic, in this work we take the recent Bottom-up Top-down architecture \cite{anderson_2017} as our base model. 
A number of works~\cite{xu2015show,fang2015captions,you2016image,wu2016encode,lu2017knowing, yang2016stacked,Lu2016} use Top-down attention mechanisms to help fine-grained and multi-stage reasoning, which is shown to be very important for vision and language tasks. 
Anderson \etal \cite{anderson_2017} propose a variant of the traditional attention mechanism, where instead of attending over convolutional features they show that attending over objects and other salient image regions gives significant improvements in VQA and captioning performance. We briefly describe this architecture below, see \cite{anderson_2017} for full details.

\begin{figure*}[t]
\centering
\includegraphics[width=0.95\textwidth]{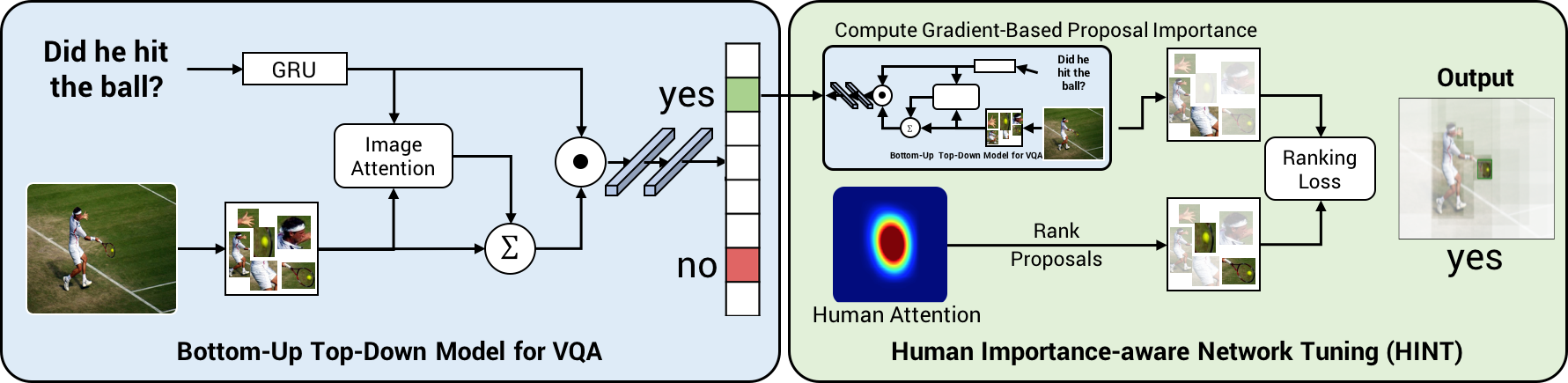}\\
\caption{Our Human Importance-aware Network Tuning (HINT) approach: Given an image and a question like ``Did he hit the ball?", we pass them through the Bottom-up Top-down architecture shown in the left. For the example shown, the model incorrectly answers `no'. We determine the proposals important for the ground-truth answer `yes' through a gradient-based importance measure. We rank the proposals through human attention and provide a ranking loss in order to align the network's importance with human importance. Tuning the model through HINT makes the model not only answer correctly, but also look at the right regions, as shown in the right.}\label{fig:hint_approach_vqa}
\vspace{-10pt}
\end{figure*}

\xhdr{Bottom-Up Top-Down Attention for VQA.} As shown in left half of Fig. \ref{fig:hint_approach_vqa}, given an image, the Bottom-up Top-down (UpDown) attention model takes as input up to $k$ image features, each encoding a salient image region. 
These regions and their features are proposals extracted from Faster-RCNN~\cite{girshick2015fast}. The question is encoded using a GRU~\cite{cho2014learning} and a soft-attention over each of the $k$ proposal features is computed using the question embedding. The final pooled attention feature is combined with the question feature using a few fully-connected layers which predict the answer. %

\xhdr{Bottom-Up Top-Down Attention for Image Captioning.} 
The image captioning model consists of two Long Short-Term Memory (LSTM) networks -- an attention LSTM and a language LSTM.
The first LSTM layer is a top-down visual attention model whose input at each time step consists of the previous hidden state of the language LSTM, concatenated with the mean-pooled bottom-up proposal features (similar to above) and an encoding of the previously generated word. 
The output of the attention LSTM does a soft attention over the proposal features.
The second LSTM is a language generation LSTM that takes as input the attended features concatenated with the output of the attention LSTM. 
The language LSTM provides a distribution over the vocabulary of words for the next time step.

\reducedSection{Human Importance-aware Network Tuning}

In this section, we describe our approach for training deep networks to rely on the same regions as humans which we call Human Importance-aware Network Tuning (HINT). %
In summary, HINT estimates the importance of input regions through gradient-based explanations and tunes the network parameters so as to align this with the regions deemed important by humans. %
We use the generic term `prediction' to refer to both answers in the case of VQA and the words generated at each time step in image captioning.

\reducedSubSection{Human Importance} \label{sec:human_importance}

In this step, we align the expert knowledge obtained from humans attention maps into a form corresponding to the network inputs. 
The Bottom-up Top-down model~\cite{anderson_2017} takes in as input region proposals. %
For a given instance, we compute an importance score for each of the proposals based on normalized human attention map energy inside the proposal box relative to the normalized energy outside the box. 

More concretely, consider a human importance map $A^d \in \mathbb{R}^{h\times w}$ that indicates the spatial regions of support for an output $d$\footnote{For VQA, these maps will vary across questions for a given image.} -- a high value $A^d[i,j]$ indicates high support for $d$ at location (i,j).  Given a proposal region $r$ with area $a_r$, we can write the normalized importance inside and outside $r$ for decision $d$ as 
$$E^d_i(r){=}\frac{1}{a_r}\sum_{ (i,j) \in r}A^d_{ij} \mbox{~~and~~} E_o^d(r){=} \frac{1}{h.w{-}a_r}\sum_{ (i,j) \notin r}A^d_{ij}$$ 
respectively. We compute the overall importance score for proposal $k$ for decision $d$ as: 

\begin{equation}
s_k^d = \frac{E_i^d(k)}{E_i^d(k) + E_o^d(k)}
\label{eq:human_importance}
\end{equation}

\vspace{5pt}
\noindent
\textbf{Human attention for VQA and captioning.}~
For VQA, we use the human attention maps collected by Das \etal ~\cite{vqahat} for a subset of the VQA~\cite{antol2015vqa} dataset. 
HAT maps are available for a total of 40554 image-question pairs -- \emph{or approximately only $\sim$6\% of the VQA dataset.}
While human attention maps do not exist for image captioning, COCO dataset \cite{LinECCV14coco} has segmentation annotations for 80 everyday occurring categories. 
We use a word-to-object mapping that links fine-grained labels like [``child'', ``man'', ``woman'', ...] to object categories
like $<$person$>$ similar to \cite{lu2018Neural}. 
We map a total of 830 visual words existing in COCO captions to 80 COCO categories.
We then use the segmentation annotations for the 80 categories as human attention for this subset of matching words. 
To be consistent with the VQA setup, we only use 6\% of the segmentation annotations.%

\reducedSubSection{Network Importance}

We define Network Importance as the importance that the given trained network places on spatial regions of the input when making a particular prediction. 
Selvaraju \etal \cite{gradcam_ijcv} proposed an approach to compute the importance of last convolutional layer's neurons. %
In their work, they focus on the last convolutional layer neurons as they serve as the best compromise between high level semantics and detailed spatial information. 
Since proposals usually look at objects and salient/semantic regions of interest while providing a good spatial resolution, we extend \cite{gradcam_ijcv} to compute importance over proposals. 
In order to obtain the importance of a proposal $r$ for ground-truth decision, $\alpha{}_{gt}^r$, we one-hot encode the score for the ground-truth output (answer in VQA and the visual word in case of captioning) $o_{gt}$ and compute its gradients \wrt proposal features as, 
\vspace{-5pt}
\begin{equation}\label{eq:network_importance}
    \alpha{}_{gt}^r =
    \overbrace{
        \sum_{i=1}^{|P|}
    }^{\text{global pooling}} \mkern-65mu
    \hspace{10pt}
    \underbrace{         
        \frac{\partial o_{gt}}{\partial P_{i}^{r}}
    }_{\text{gradients via backprop}}
\end{equation}

\noindent %
Note that we compute the importance for the ground-truth decision, and not predicted. 
Human attention for incorrect decisions are not available and are conceptually ill-posed because it is difficult to define what correct `evidence' for an incorrect prediction would be. %

\reducedSubSection{Human-Network Importance Alignment}

At this stage, we now have two sets of importance scores -- one computed from the human attention and another from network importance -- that we would like to align. Each set of scores is calibrated within itself; however, absolute values are not comparable between the two as human importance lies in $[0,1]$ while network importance is unbounded. Consequentially, we focus on the relative rankings of the proposals, applying a ranking loss -- specifically, a variant of Weighted Approximate Rank Pairwise (WARP) loss.

\xhdr{Ranking loss.}~ 
At a high level, our ranking loss searches all possible pairs of proposals and finds those pairs where the pair-wise ranking based on network importance disagrees with the ranking from human importance. 
Let $\mathcal{S}$ denote the set of all such misranked pairs. 
For each pair in $\mathcal{S}$, the loss is updated with the absolute difference between the network importance score for the proposals pair. %

\begin{equation}\label{eq:WARP_loss}
   \mathcal{L} = \sum_{(r',r)\in\mathcal{S}} \left| \alpha{}^{r'}_{-} - \alpha{}^{r}_{+} \right|
 \end{equation}
 
 \noindent where $r$ and $r'$ are the proposals whose order based on neuron importance does not align with human importance and $+$ indicates that proposal $r$ is more important compared to $r'$ according to human importance. 
 
 \xhdr{Importance of task loss. } In order to retain performance at the base task, it is necessary to include the original task loss $\lambda L_{Task}$ -- cross-entropy for VQA and negative log-likelihood in case of image captioning. To trade-off between the two, we introduce a multiplier $\lambda$ such that the final HINT loss becomes, 
 
 \begin{equation}\label{eq:loss}
   \mathcal{L_{HINT}} =  \sum_{(r',r)\in\mathcal{S}}\left| \alpha{}^{r'}_{-} - \alpha{}^{r}_{+} \right| + \lambda L_{Task}
 \end{equation}
 
 \noindent The first term encourages the network to base predictions on the correct regions and the second term encourages it to actually make the right prediction.

Note that network importances $\alpha{}$ are gradients of the score with respect to proposal embeddings. Thus they are a function of all the intermediate parameters of the network ranging from the model attention layer weights to the final fully-connected layer weights. 
 Hence an update through an optimization algorithm (gradient-descent or Adam) with the given loss in \eqref{eq:loss} requires computation of second-order gradients, and would affect all the network parameters. We use PyTorch \cite{paszke2017automatic} which has this functionality.

\reducedSection{Experiments and Analysis}

\begin{figure*}[t]

 \centering
 \begin{subfigure}{.5\textwidth}
  \centering
  \includegraphics[scale=0.5]{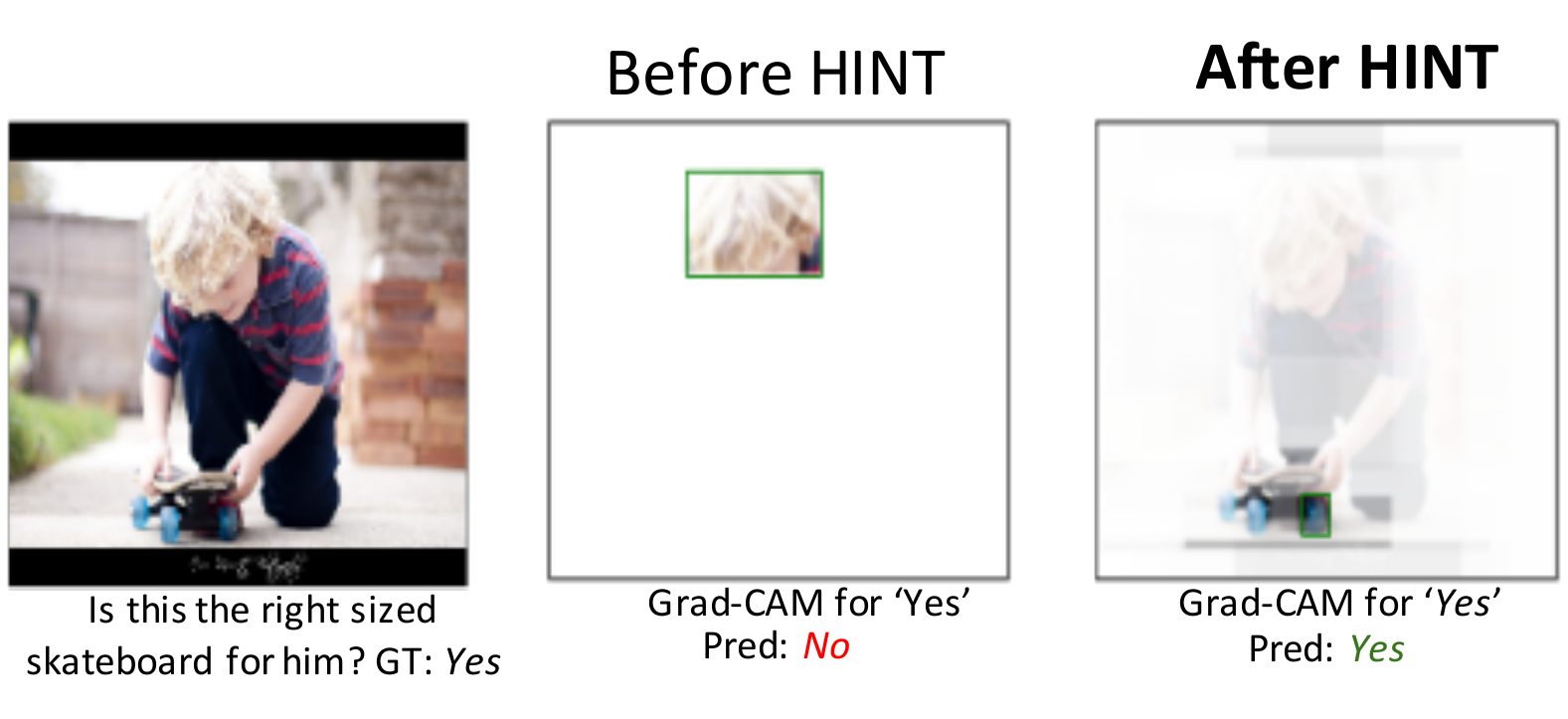}
  \vspace{-10pt}
  \caption{}
\end{subfigure}%
 \vrule\ \hspace{-0.05in}%
\begin{subfigure}{.5\textwidth}
  \centering
  \includegraphics[scale=0.5]{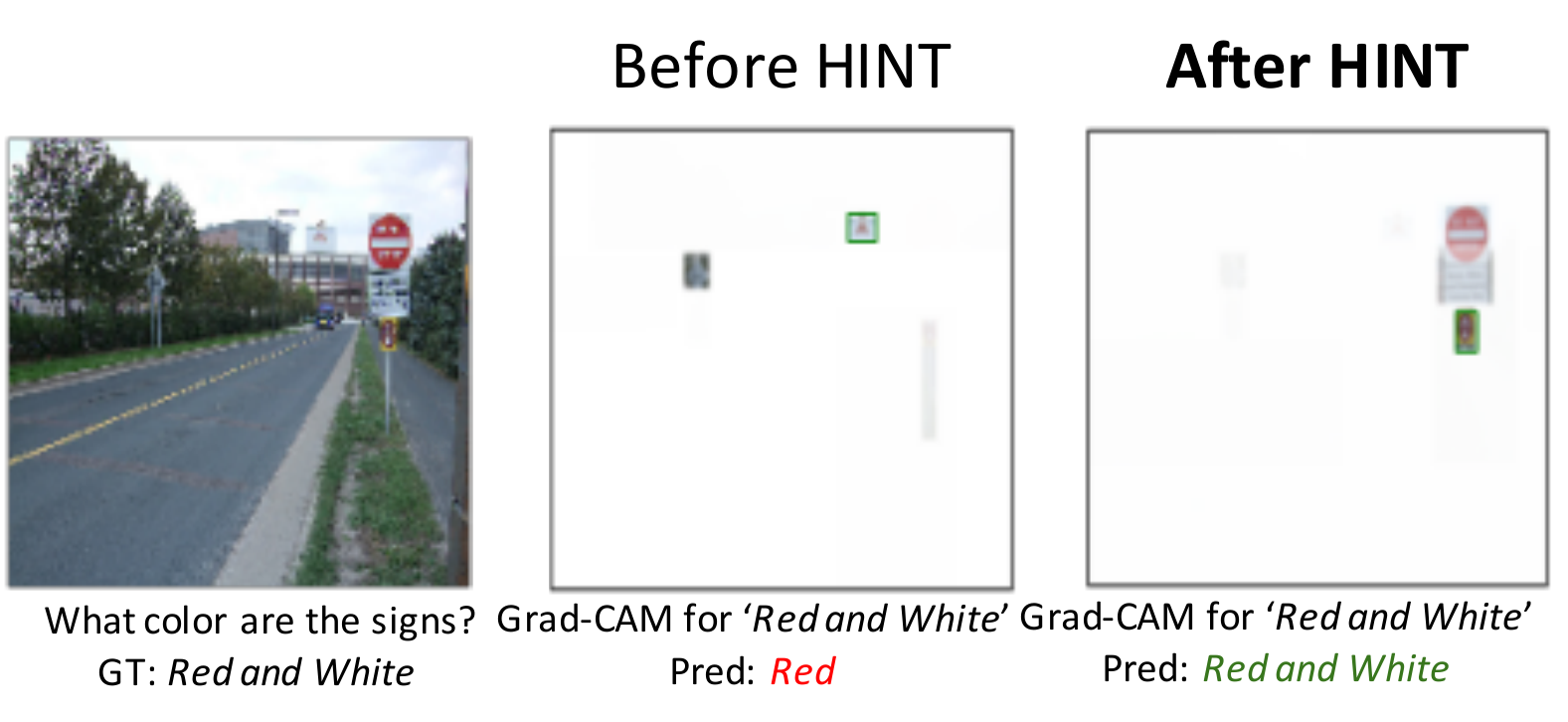}
  \vspace{-10pt}
  \caption{}
\end{subfigure}
\begin{subfigure}{.5\textwidth}
  \centering
  \includegraphics[scale=0.5]{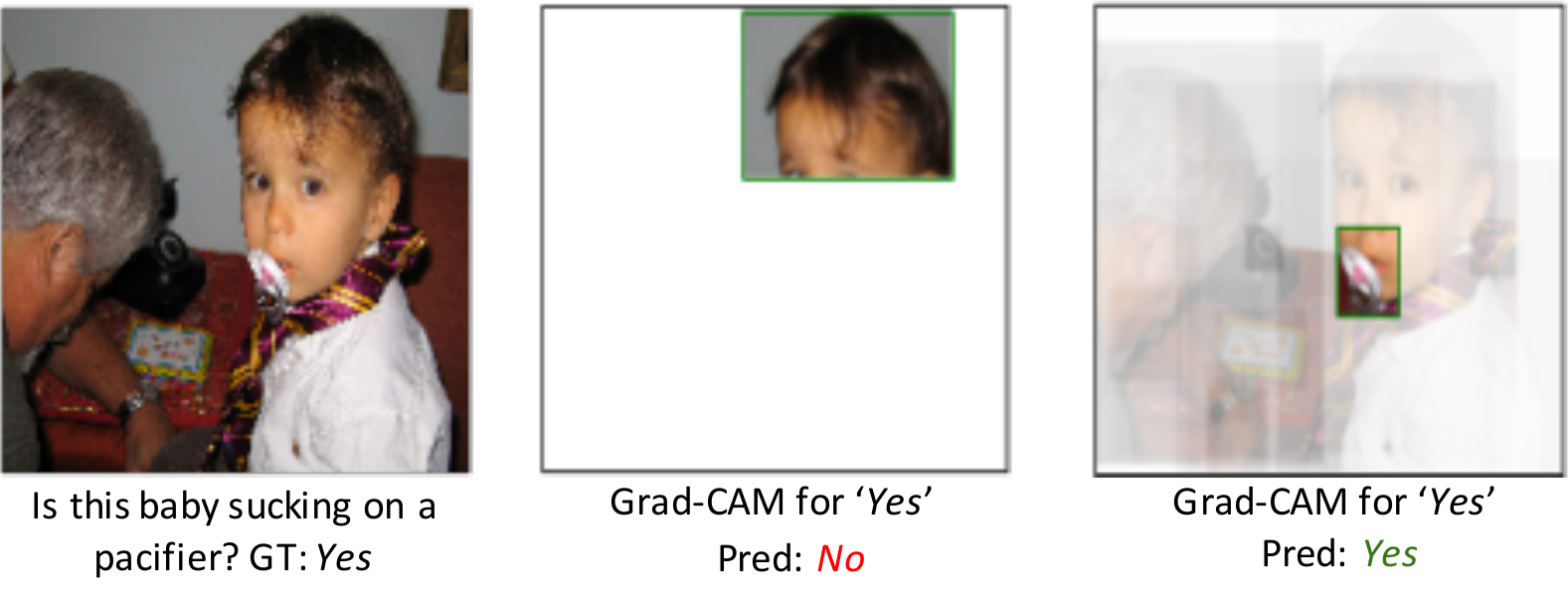}
  \vspace{-10pt}
  \caption{}
\end{subfigure}%
\vrule\ \hspace{-0.05in}%
\begin{subfigure}{.5\textwidth}
  \centering
  \includegraphics[scale=0.5]{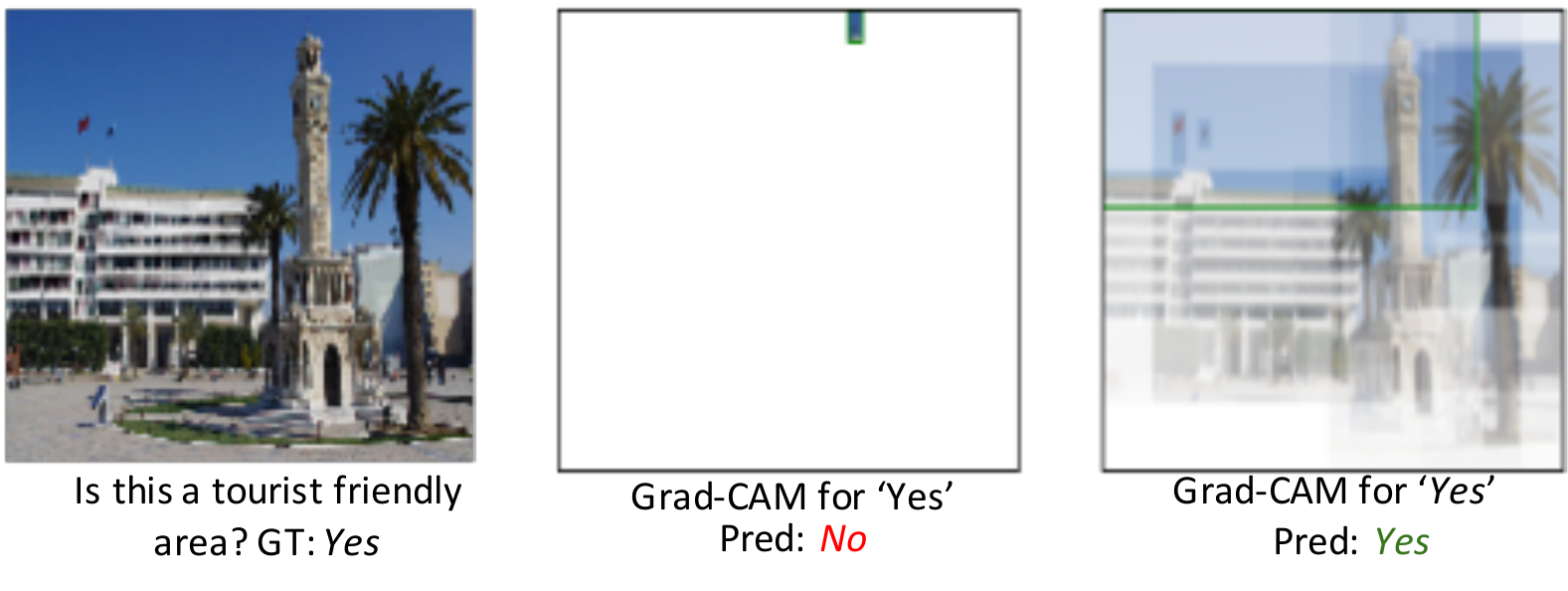}
  \vspace{-10pt}
  \caption{}
\end{subfigure}
\vspace{-10pt}
\caption{Qualitative comparison of models on validation set before and after applying HINT. For each example, the left column shows the input image along with the question and the ground-truth (GT) answer from the VQA-CP val split. In the middle column, for the base model we show the explanation visualization for the GT answer along with the model's answer. Similarly we show the explanations and predicted answer for the HINTed models in the third column. We see that the HINTed model looks at more appropriate regions and answers more accurately. For example, for the example in (a), the base model only looks at the boy, and after we apply HINT, it looks at both the boy and the skateboard in order to answer `Yes'. After applying HINT, the model also changes its answer from `No' to `Yes'. More qualitative examples can be found in the supplementary material.}
\label{fig:qual_vqa}
\vspace{-8pt}
\end{figure*}

In this section we describe the experimental evaluation of our approach on VQA and Image Captioning. 

\xhdr{VQA.} For VQA, we evaluate on the standard VQA split and the VQA-CP \cite{agrawal2018don} split. Recall from Section \ref{sec:rel} that VQA-CP is a restructuring of VQAv2 \cite{goyal2016making} that is designed such that the answer distribution in the training set differs significantly from that of the test set. For example, while the most popular answer in train for ``What sport ...'' questions might be ``tennis'', in test it might be ``volleyball''. Without proper visual grounding, models trained on this dataset will generalize poorly to the test distribution. In fact, \cite{agrawal2018don} and \cite{advregvqa_nips_2018} report significant performance drops for state-of-the-art VQA models on this challenging, language-bias sensitive split. For our experiments, we pretrain our Bottom-Up Top-Down model on respective training splits before fine-tuning with the HINT loss. 
Recall that our approach includes the task loss; 
We use $\lambda_{vqa}=10$ for our experiments.

\begin{table*}[t]
\centering
\resizebox{0.9\textwidth}{!}{
\centering
\renewcommand{\arraystretch}{1.11}
\setlength{\tabcolsep}{8pt}
\begin{tabular}{l p{0pt}  p{5pt} c c c c  p{5pt}  c c c c}
\toprule
\multirow{2}{*}{Model} && &  \multicolumn{4}{c}{VQA-CP \texttt{test}} & & \multicolumn{4}{c}{VQAv2 \texttt{val}}\\
 && & \footnotesize Overall & \footnotesize Yes/No & \footnotesize Number & \footnotesize Other & &\footnotesize Overall &\footnotesize Yes/No &\footnotesize Number &\footnotesize Other\\
\midrule
SAN \cite{yang2016stacked} && &  24.96 & 38.35  & 11.14  & 21.74 & & 52.41 & 70.06 & 39.28 & 47.84\\
UpDn \cite{anderson_2017} && & 39.49 & 45.21 & 11.:96 & 42.98 & & 62.85 & 80.89 &  42.78 & 54.44\\
\midrule
GVQA \cite{agrawal2018don}$^\dagger$ &&  &  31.30 & 57.99   & 13.68  & 22.14 & &  48.24 & 72.03   & 31.17  & 34.65\\
\midrule
UpDn + Attn. Align  && & 39.37 & 43.02 &11.89 & 45.00 && 63.24 & 80.99 & 42.55 & 55.22\\
UpDn + AdvReg \cite{advregvqa_nips_2018}$^\dagger$ && & 41.17 & 65.49 & \textbf{15.48} & 35.48 & & 62.75 & 79.84 & 42.35 & 55.16\\
UpDn + HINT \scriptsize{(ours)}  && & \textbf{46.73} & \textbf{67.27} & 10.61 & \textbf{45.88} & & \textbf{63.38} & \textbf{81.18} & \textbf{42.99} & \textbf{55.56}\\
\bottomrule
\end{tabular}}
\vspace{-7pt}
\caption{Results on compositional (VQA-CP) and standard split (VQAv2). We see that our approach (HINT) gets a significant boost of over 7\% from the base UpDn model on VQA-CP and minor gains on VQAv2. The Attn.~Align baseline sees similar gains on VQAv2, but fails to improve grounding on VQA-CP. 
Note that for VQAv2, during HINT finetuning we apply the VQA cross entropy loss even for the samples without human attention annotation. 
$\dagger$ results taken from corresponding papers.}
\label{tab:hint_vqa}
\vspace{-19pt}
\end{table*}

We compare our approach against strong baselines and existing approaches, specifically:
\begin{itemize}[nosep,leftmargin=1em,labelwidth=*,align=left]
\item \textbf{Base Model (UpDn)} We compare to the base Bottom-up Top-down model without our HINT loss.
\item \textbf{Attention Alignment (Attn.~Align.)} We replace gradient supervision with attention supervision keeping everything else the same. 
The Bottom-up Top-down model uses soft attention over object proposals -- essentially predicting a set of attention scores for object proposals based on their relevancy to the question. These attention scores are much like the network importances we compute in HINT; however, they are functions only of the network prior to attention prediction. We apply the HINT ranking loss between these attention weights and human importances as computed in Equation \eqref{eq:human_importance}. 
\item \textbf{Grounded VQA (GVQA).} As discussed in Section \ref{sec:rel}, \cite{agrawal2018don} introduced a grounded VQA model that explicitly disentangles vision and language components and was developed alongside the VQA-CP dataset.
\item \textbf{Adversarial Regularization (AdvReg).} \cite{advregvqa_nips_2018} introduced an adversarial regularizer to reduce the effect of language-bias in VQA by explicitly modifying question representations to fool a question-only adversary model.
\end{itemize}

\xhdr{Image Captioning.} For captioning, we evaluate on the standard `Karpathy' split and the robust captioning split introduced by Lu \etal in \cite{lu2018Neural}. 
The robust split has varying distribution of co-occurring objects between train and test. 
We pretrain our Bottom-up Top-down captioning model on the respective training splits and apply our approach, HINT. 
Note that the HINT loss is applied only for the time steps corresponding to the 830 visual words in the caption that we obtain in Section \ref{sec:human_importance}.

\begin{figure*}[ht!]
 \centering
\begin{subfigure}{.48\textwidth}
  \centering
  \hspace{-10pt}
  \includegraphics[scale=0.5]{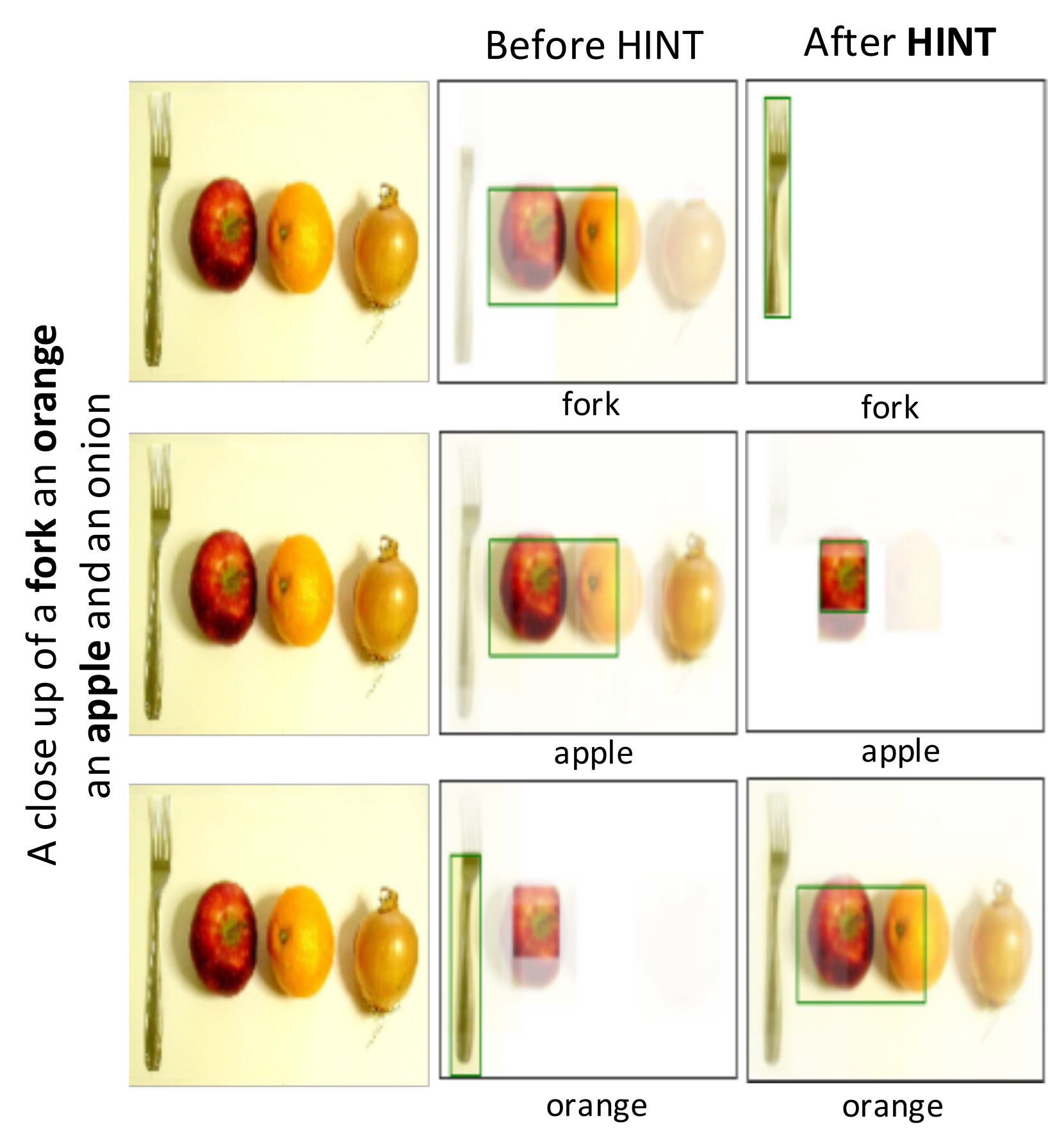}
  \vspace{-10pt}
  \caption{}
\end{subfigure}%
\vrule\ \hspace{-0.07in}%
\begin{subfigure}{.48\textwidth}
  \centering
  \includegraphics[scale=0.5]{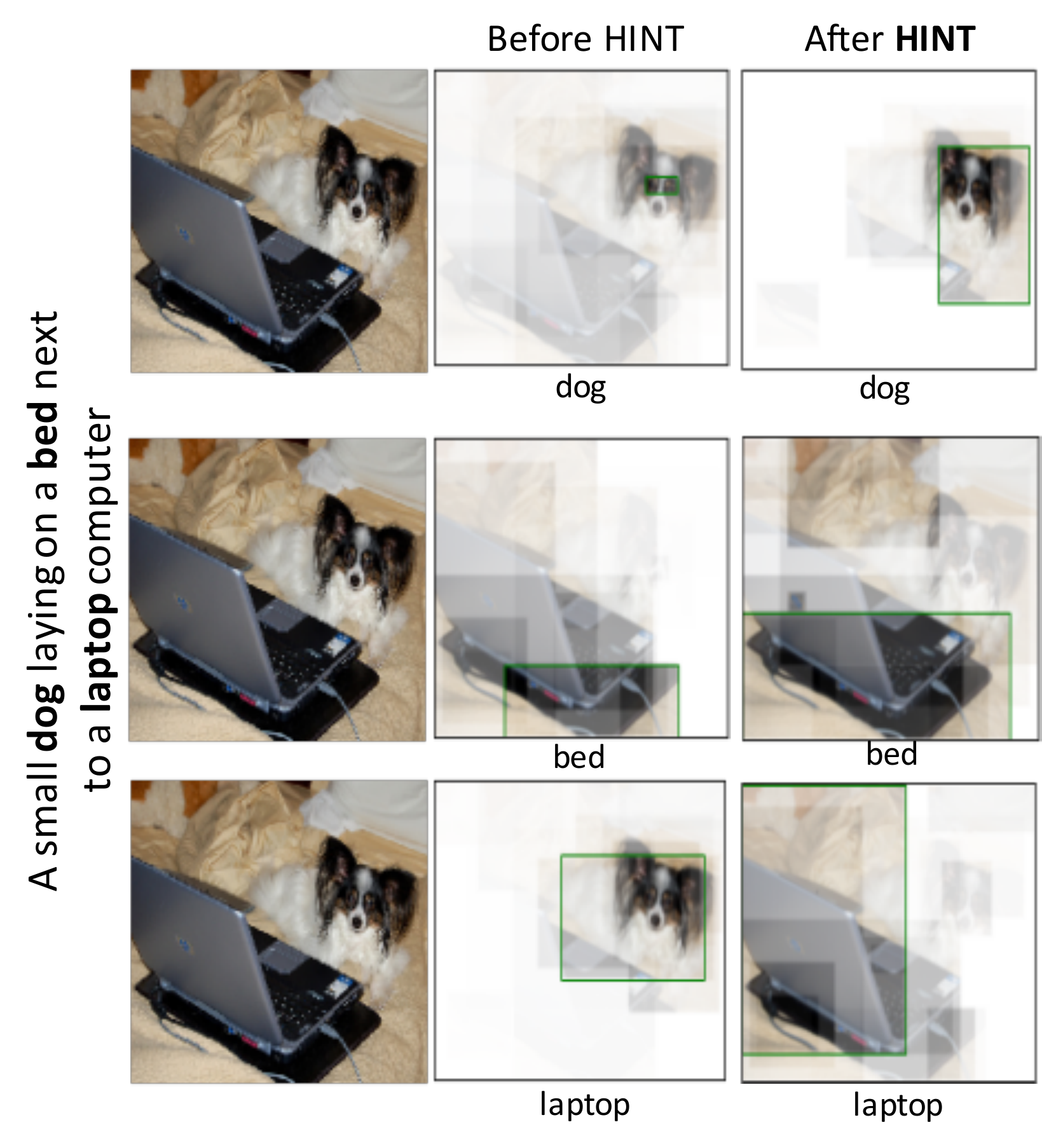}
  \vspace{-20pt}
  \caption{}
\end{subfigure}
\begin{subfigure}{.48\textwidth}
  \centering
  \hspace{-10pt}
  \includegraphics[scale=0.20]{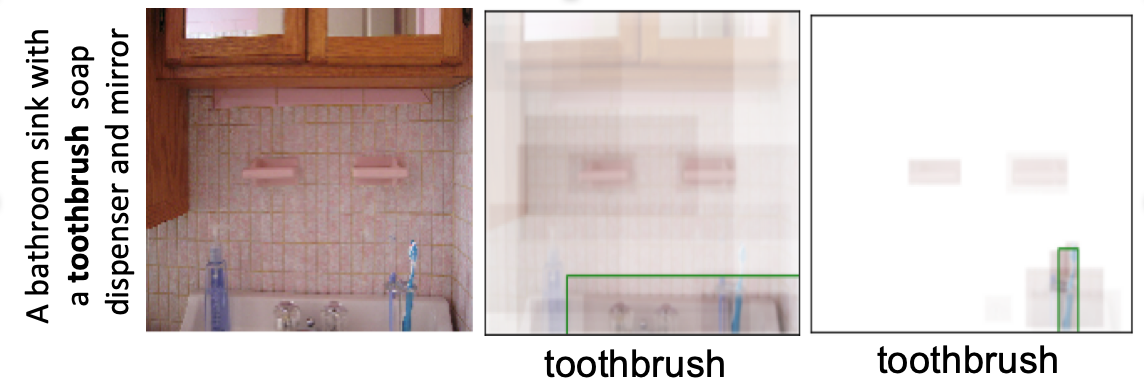}
  \caption{}
\end{subfigure}%
\vrule\ \hspace{-0.1in}%
\begin{subfigure}{.48\textwidth}
  \centering
  \includegraphics[scale=0.20]{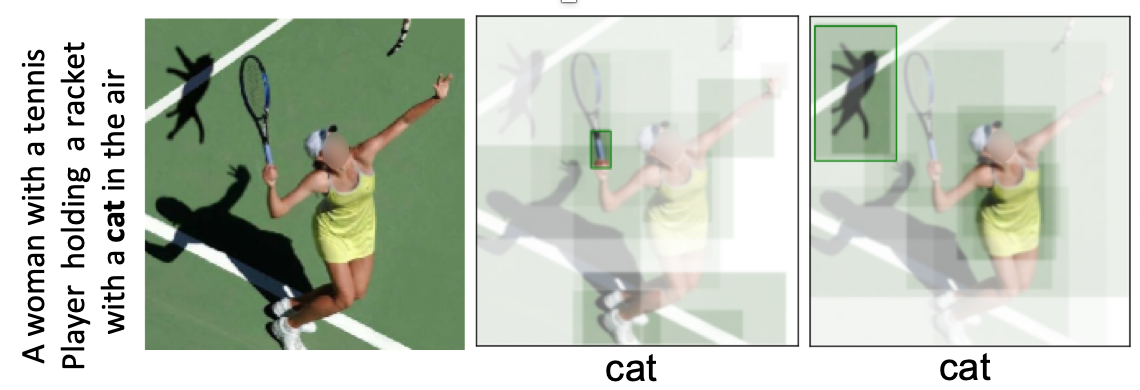}
  \caption{}
\end{subfigure}
\vspace{-5pt}
\caption{Qualitative comparison of captioning models on validation set before and after applying HINT. For each example, the left column shows the input image along with the ground-truth caption from the COCO robust split. In the middle column, for the base model we show the explanation visualization for the visual word mentioned below. Similarly we show the explanations for the HINTed models in the third column. We see that the HINTed model looks at more appropriate regions. For example in (a) note how the HINTed model correctly localizes the fork, apple and the orange when generating the corresponding visual words, but the base model fails to do so. Interestingly the model is able to ground even the shadow of a cat in (f)! More qualitative examples can be found in the supplementary material.
}
\vspace{-10pt}\label{fig:qual_captioning}
\end{figure*}

\reducedSubSection{HINT for Visual Question Answering}\label{sec:vqa}

Table \ref{tab:hint_vqa} shows results for our models and prior work on VQA-CP test and VQAv2 val. We summarize key results:

\xhdr{HINT reduces language-bias.} For VQA-CP, our HINTed UpDown model significantly improves over its base architecture alone by ~7
percentage point gain in overall accuracy. Further, it outperforms existing approaches based on the same UpDn architecture (41.17 for AdvReg vs 46.73 for HINT), setting a new state-of-the-art for this problem. We do note that our approach uses additional supervision in the form of human attention maps for ~6\% of training images.%

\xhdr{HINT improves grounding without reducing standard VQA performance.} Unlike previous approaches for language-bias reduction which cite trade-offs in performance between the VQA and VQA-CP splits \cite{advregvqa_nips_2018, agrawal2018don}, we find our HINTed UpDn model actually improves on standard VQA -- making HINT the first ever approach to show simultaneous improvement on both the standard and compositional splits.

\xhdr{Attn.~Align is ineffective compared to HINT.} 
A surprising (to us at least) finding and motivating observation of this work is that directly supervising model attention (as in Attn.~Align) is ineffective at reducing language-bias and improving visual grounding as measured by VQA-CP, begging the question -- {why does our gradient supervision succeed where attention supervision fails?}

\label{sec:faithfulness}

We argue this results from gradient-based explanations being 1) a function of all network parameters unlike attention alignment and 2) more faithful to model decisions than model attention. As we've discussed previously, attention is a bottom-up computation and supervising it cannot directly affect later network layers, whereas our HINT approach does. 
To assess faithfulness, we run occlusion studies similar to those in \cite{gradcam_ijcv,zeiler_eccv14}. 
We measure the difference in model scores for the predicted answer when different proposal features for the image are masked and forward propagated, taking this delta as an importance score for each proposal. 
We find that rank correlation between model attention and occlusion-based importance is only $0.10$, compared to $0.48$ for gradient-based importance -- demonstrating our claim that \emph{model attention only loosely relates to how the model actually arrives at its decision. } %
As such, attention alignment simply requires the model to predict human-like attention, not necessarily to care about them when making decisions. 
On the other hand, HINT aligns gradient-based importance with respect to model decisions, ensuring that human specified regions are actually used by the network -- resulting in a model that is \emph{right for the right reasons}.

\xhdr{Varying the amount of human attention supervision. }
\begin{wrapfigure}{r}{0.40\columnwidth} 
\vspace{-20pt}
\begin{center}
\hspace{-20pt}\includegraphics[width=1.15\linewidth]{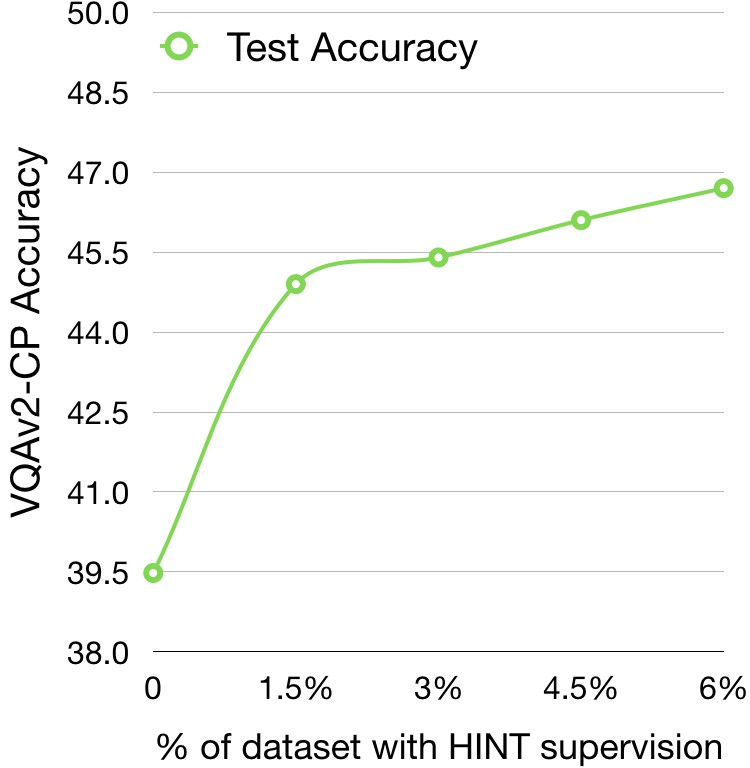}
\end{center}
\label{fig:vary_hint}
\vspace{-25pt}
\end{wrapfigure}
The plot to the right shows performance for different amounts of Human Attention maps for VQA-CP.
Note that the x-axis goes from using no HINT supervision to using all the Human attention maps during training, which amounts to $6\%$ of the VQAv2 data. 
Note that with human attention supervision for just 1.5\% of the VQA dataset, our approach achieves a 5 \% improvement in performance.

\xhdr{Qualitative examples.~}      Fig.~\ref{fig:qual_vqa} shows qualitative examples showing the effect of applying HINT to the Bottom-up Top-down VQA model. 
Fig.~\ref{fig:qual_vqa} (b) shows an image and a question, \myquote{What color are the signs?}, the base model answers ``Red" which is partially correct, but it fails to ground the answer correctly. The HINTed model not only answers ``Red and White" correctly but also looks at the red stop sign and the white street sign.

\reducedSubSection{HINT for Image Captioning}\label{sec:captioning}

Our implementation of the Bottom-up Top-down captioning model in Pytorch~\cite{paszke2017automatic} achieves a CIDEr \cite{DBLP:journals/corr/VedantamZP14a} score of 1.06 on the standard split and 0.90 on the robust split.
Upon applying HINT to the base model trained on the robust split, we obtain a CIDEr score of 0.92, an improvement of 0.02 over the base model.  
For the model trained on the standard split, performance drops by 0.02 in CIDEr score (1.04 compared to 1.06). As we show in the following sections, the lack of improvement in score does not imply a lack of change -- we find the model shows significant improvements at grounding, which we evaluate in Section \ref{sec:eval_grounding}. 
Note that our setup for captioning does not require \emph{task-specific human attention}, and instead allows us to directly leverage existing annotations which were collected for a different task (image segmentation). 

\vspace{5pt}
\noindent
\textbf{Qualitative examples.~}  Fig.~\ref{fig:qual_captioning} shows qualitative examples that indicate significant improvements in grounding performance of HINTed models. For example Fig.~\ref{fig:qual_captioning} (a) shows how a model trained with HINT is able to simultaneously improve grounding for the 3 visual words present in the ground-truth caption. 
We see that HINT also helps with making models focus on individual object occurrences rather than using context, as shown in Fig.~\ref{fig:qual_captioning} (c, d, e, f).

\reducedSection{Evaluating Grounding}\label{sec:eval_grounding}
In Sections \ref{sec:vqa} and \ref{sec:captioning} we evaluated the effect of HINT on the task performance, with generalization to robust dataset splits serving as an indirect evaluation of grounding. %
In this section we directly evaluate the grounding ability of models tuned with HINT.

\reducedSubSection{Correlation with Human Attention}

In order to evaluate the grounding ability of models before and after applying HINT, we compare the network importances for the ground-truth decision (as in Equation \eqref{eq:network_importance}) with the human attention as computed in Equation \eqref{eq:human_importance} for both the base model and the model fine-tuned with HINT. 
We then compute the rank correlation between the network importance scores and human importance scores for images from the VQA-CP and COCO robust test splits. 
We report Spearman's rank correlation between explanations from the base model and the HINTed model. 

\xhdr{VQA.}
For the model trained on VQA-v2, we find that the Grad-CAM based attention for base model obtains a Spearman's rank correlation of -0.09 with human attention maps \cite{vqahat}. Note that the range of rank-correlation is -1 to 1, so near 0 indicates no correlation. We find that the HINTed model obtains a correlation of 0.18. 

\begin{figure}[t]
\centering
\includegraphics[width=0.5\textwidth]{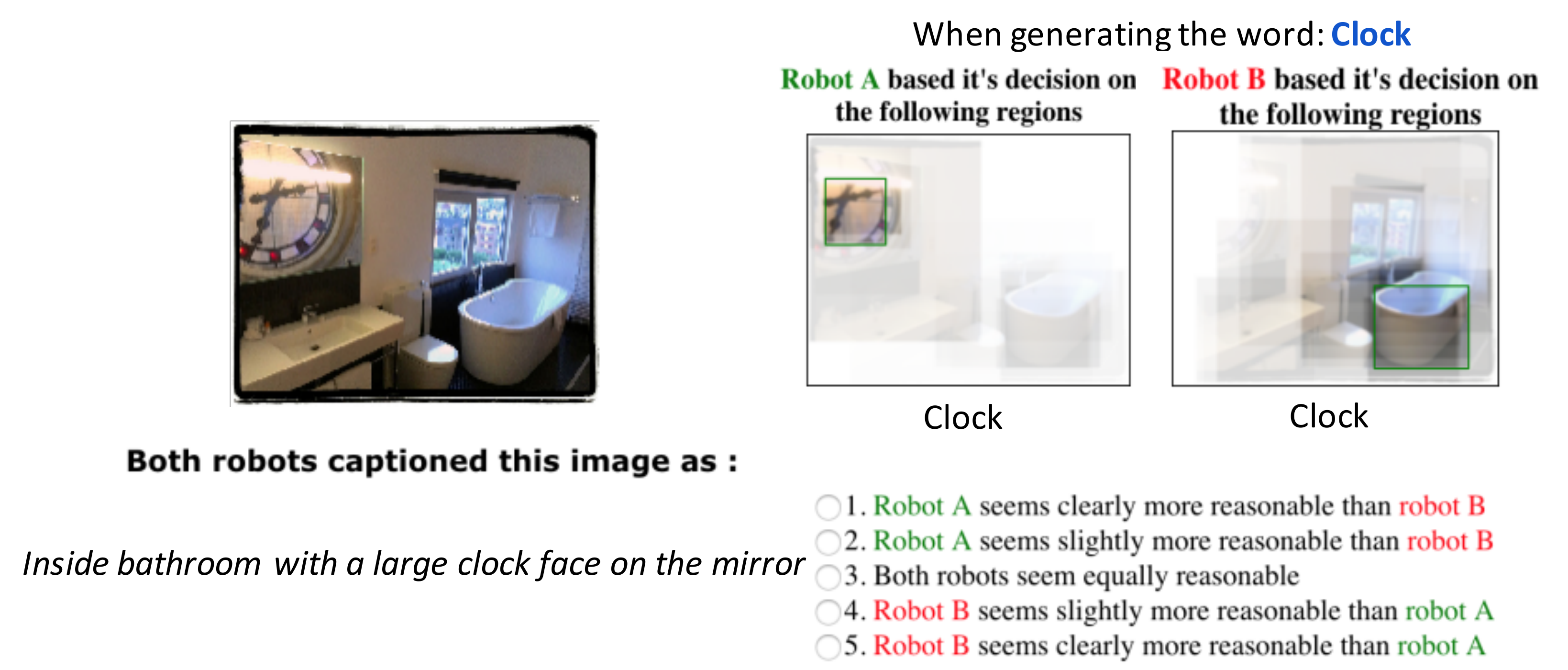}\\
\vspace{-3pt}
\caption{AMT interface for evaluating the baseline captioning model and our HINTed model. HINTed model outperforms baseline model in terms of human trust.}
\label{fig:captioning_amt}
\vspace{-15pt}
\end{figure}

\xhdr{Image Captioning.}
For the model trained on the COCO robust split, the Grad-CAM based attention for base model achieves a rank correlation of 0.008 with COCO segmentation maps for the visual words, and the model after HINTing achieves a correlation of 0.17. 

This rank correlation measure matches the intent of the rank-based HINT loss, but this result shows that the visual grounding learned during training generalizes to new images and language contexts better than the baseline model.

\reducedSection{Evaluating Trust}\label{sec:human_evaluation}

In the previous section we evaluate if HINTed models attend to the same regions as humans when forced into making predictions. 
Having established that, we turn to understanding whether this improved grounding translates to increased human trust in HINTed models. We focus this study on our image captioning models.

We conduct human studies to evaluate if based on individual prediction explanations from two models -- the base model and one with improved grounding through HINT -- humans find either of the models more trustworthy. In order to tease apart the effect of grounding from the accuracy of the models being visualized, we only visualize predictions corresponding to the ground-truth caption for both models.

For a given ground truth caption, we show study participants the network importance explanation for a ground truth visual word as well as the whole caption. Workers were then asked to rate the reasonableness of the models relative to each other on a 5-point Likert scale of clearly more/less reasonable (+/-$2$), slightly more/less reasonable (+/-$1$), and equally reasonable ($0$). This interface is shown in Fig.~\ref{fig:captioning_amt}. 
In order to eliminate any biases, the base and HINTed models were assigned to be `model1' with equal probability.

In total, $42$ Amazon Mechanical Turk (AMT) workers participated in the study, producing 1000 responses (5 annotations corresponding to 200 image pairs).
In 49.9 \% of instances, participants preferred HINT compared to only 33.1 \% for the base model. 
These results indicate that HINT helps models look at appropriate regions, and that this in turn makes the model more trustworthy. 

\reducedSection{Does HINT also improve model attention?}

While HINT operates on answer gradient maps, we find it also improves feed-forward model attention. 
For VQA, we compute IoU of the top scoring proposal box with the human attention maps from Park \etal 2018. 
UpDn trained on VQA-CP obtained an IoU of 0.57 whereas after applying HINT we achieve an IoU of 0.63. 

We conduct human studies (similar to Section \ref{sec:human_evaluation}) to evaluate trust based on model attention. 
We collected 10 responses each for 100 randomly sampled image-question pairs. 
31\% of respondents found HINTed VQA-CP model to be more trustworthy compared to 16.5\% for the base model. 
This was not the primary objective of our approach but is a promising outcome for feed-forward attention!
\reducedSection{Conclusion}
We presented Human Importance-aware Network Tuning (HINT), a general framework for aligning network sensitivity to spatial input regions that humans deemed as being relevant to a task. We demonstrated this method's effectiveness at improving visual grounding in vision and language tasks such as VQA and Image Captioning. 
We also show that better grounding not only improves the generalization capability of models to changing test distributions, but also improves the trust-worthiness of model. 

Taking a broader view, the idea of regularizing network gradients to achieve desired computational properties (grounding in our case) may prove to be more widely applicable to problems outside of vision and language -- enabling users to provide focused feedback to networks. 

\small{
\xhdr{Acknowledgements.}
Georgia Tech's effort was supported in part by NSF, AFRL, DARPA, ONR YIPs, Samsung GRO, ARO PECASE. The views and conclusions contained herein are those of the authors and should not be interpreted as necessarily representing the official policies or endorsements, either expressed or implied, of the U.S. Government, or any sponsor.}

\pagebreak
\begin{appendices}

\begin{figure*}[t]
 \centering
 \begin{subfigure}{.5\textwidth}
  \centering
  \includegraphics[scale=0.25]{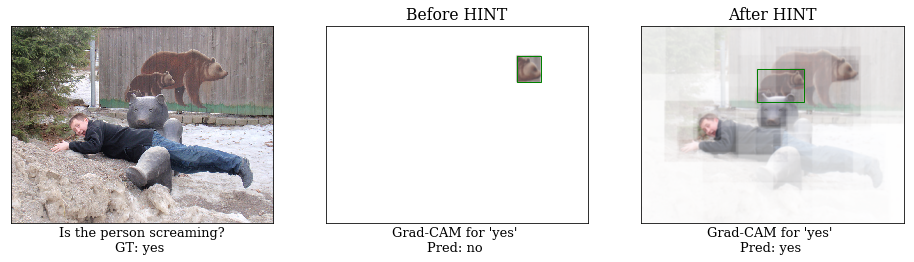}
  \vspace{-10pt}
  \caption{}
\end{subfigure}%
\begin{subfigure}{.5\textwidth}
  \centering
  \includegraphics[scale=0.25]{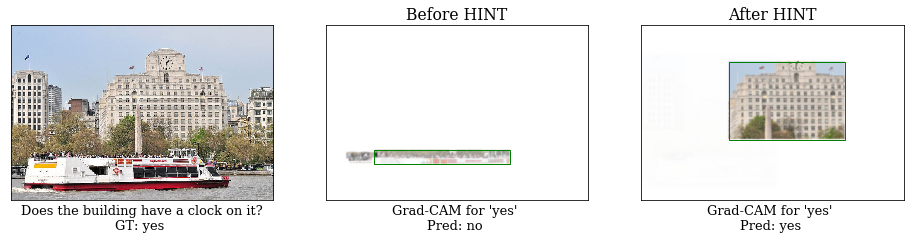}
  \vspace{-10pt}
  \caption{}
\end{subfigure}
\begin{subfigure}{.5\textwidth}
  \centering
  \includegraphics[scale=0.25]{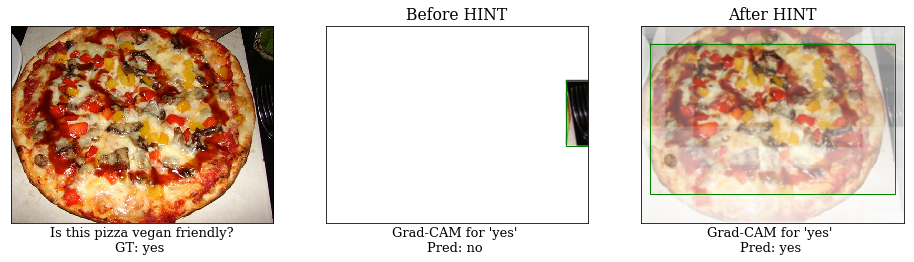}
  \vspace{-10pt}
  \caption{}
\end{subfigure}%
\begin{subfigure}{.5\textwidth}
  \centering
  \includegraphics[scale=0.25]{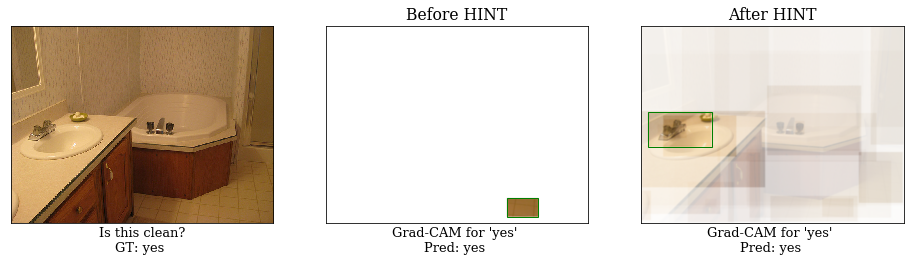}
  \vspace{-10pt}
  \caption{}
\end{subfigure}
\begin{subfigure}{.5\textwidth}
  \centering
  \includegraphics[scale=0.25]{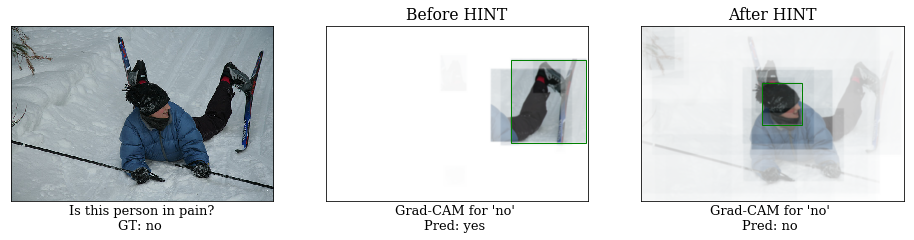}
  \vspace{-10pt}
  \caption{}
\end{subfigure}%
\begin{subfigure}{.5\textwidth}
  \centering
  \includegraphics[scale=0.25]{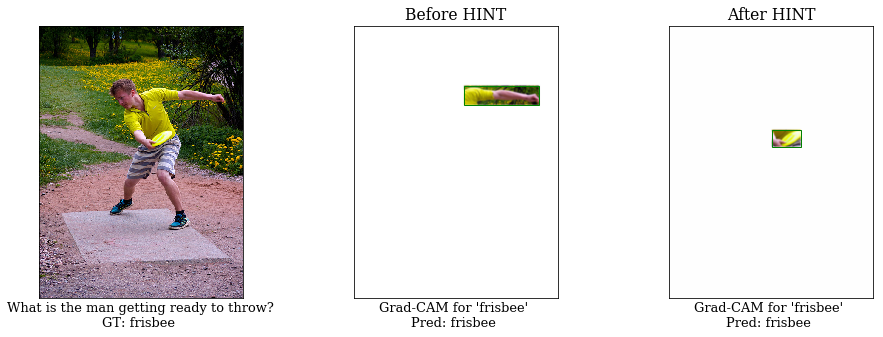}
  \vspace{-10pt}
  \caption{}
\end{subfigure}
\begin{subfigure}{.5\textwidth}
  \centering
  \includegraphics[scale=0.25]{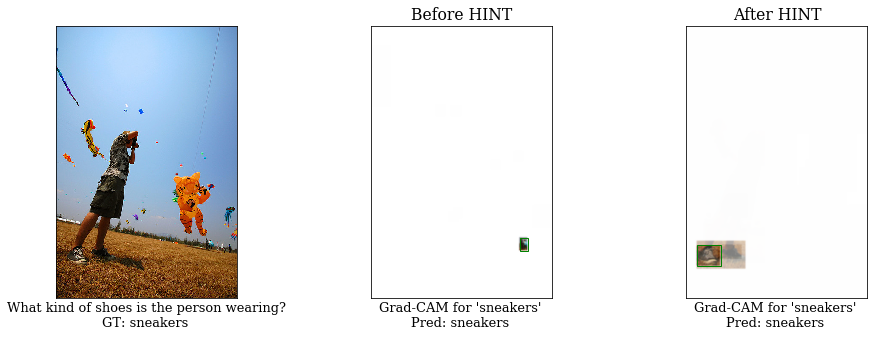}
  \vspace{-10pt}
  \caption{}
\end{subfigure}%
\begin{subfigure}{.5\textwidth}
  \centering
  \includegraphics[scale=0.25]{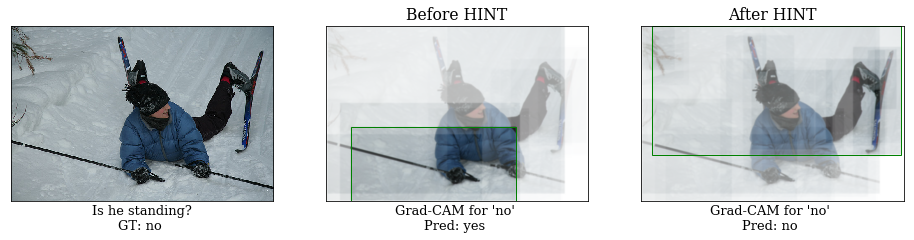}
  \vspace{-10pt}
  \caption{}
\end{subfigure}
\begin{subfigure}{.5\textwidth}
  \centering
  \includegraphics[scale=0.25]{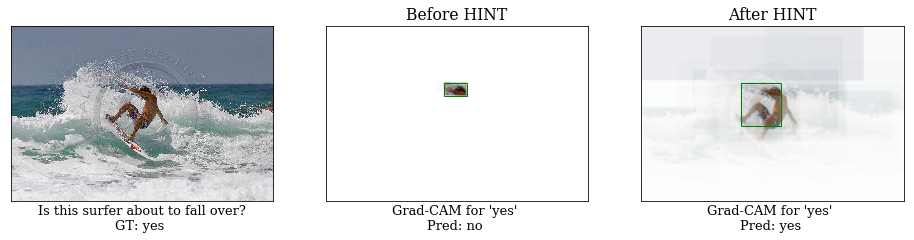}
  \vspace{-10pt}
  \caption{}
\end{subfigure}%
\begin{subfigure}{.5\textwidth}
  \centering
  \includegraphics[scale=0.25]{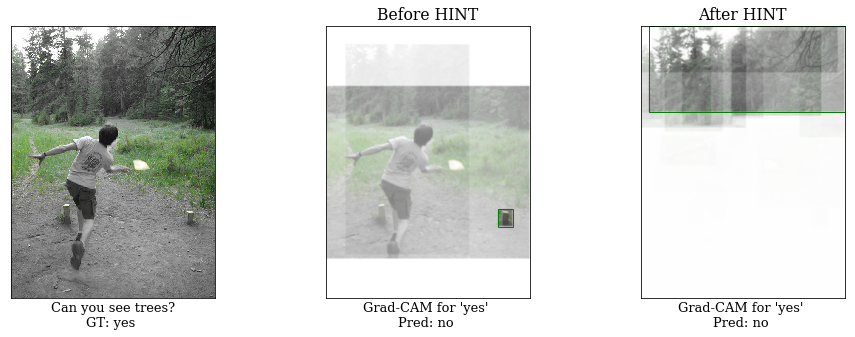}
  \vspace{-10pt}
  \caption{}
\end{subfigure}
\begin{subfigure}{.5\textwidth}
  \centering
  \includegraphics[scale=0.25]{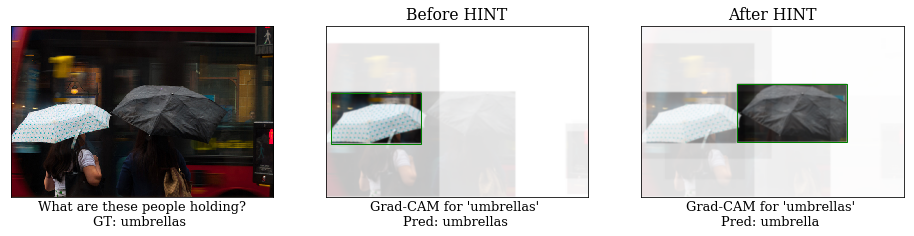}
  \vspace{-10pt}
  \caption{}
\end{subfigure}%
\begin{subfigure}{.5\textwidth}
  \centering
  \includegraphics[scale=0.25]{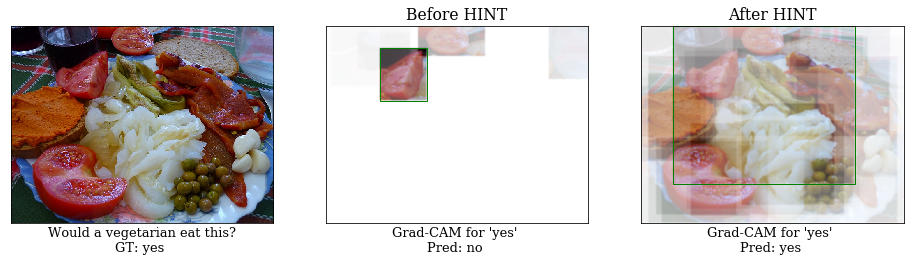}
  \vspace{-10pt}
  \caption{}
\end{subfigure}
\begin{subfigure}{.5\textwidth}
  \centering
  \includegraphics[scale=0.25]{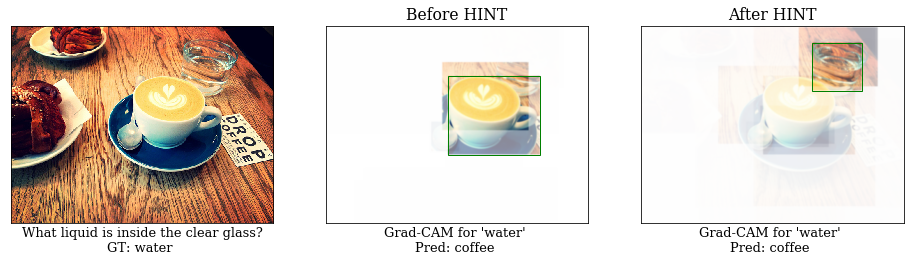}
  \vspace{-10pt}
  \caption{}
\end{subfigure}%
\begin{subfigure}{.5\textwidth}
  \centering
  \includegraphics[scale=0.25]{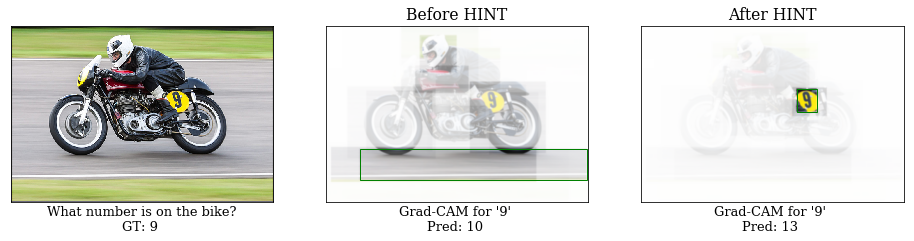}
  \vspace{-10pt}
  \caption{}
\end{subfigure}
\vspace{-10pt}
\caption{Qualitative comparison of models before and after applying HINT. The left column shows the input image along with the question and the ground-truth (GT) answer from the VQA-CP val split. In the middle column, for the base model we show the explanation visualization for the GT answer along with the model's answer. Similarly we show the explanations and predicted answer for the HINTed models in the third column. We see that the HINTed model looks at more appropriate regions and answers better. %
}
\vspace{-15pt}\label{fig:qual_vqa}
\end{figure*}
\section{Qualitative examples}

In Fig. \ref{fig:qual_vqa} we show examples applying HINT for the Bottom-up Top-down VQA model. The left column shows the input image along with the question and the ground-truth (GT) answer from the VQA-CP val split. In the middle column, for the base model we show the explanation visualization for the GT answer along with the model's answer. Similarly we show the explanations and predicted answer for the HINTed models in the third column. We see that the HINTed model not only looks at more appropriate regions compared to the base models. 

Fig. \ref{fig:qual_vqa} (a) shows an image and a question,``Is the person screaming". Not only does the base model answer ``no" incorrectly, but it also cannot localize the right answer -- looks just at the bear for ``yes". 
The HINTed model answers ``yes" correctly and looks at both the bear and the face of the person. 
For the image \ref{fig:qual_vqa} (b) with question ``Does the building have a clock on it?", the base model incorrectly answers no, whereas the HINTed model not only asnwers `yes' correctly, it also localizes the clock on the building. 
The bottom row shows two examples where HINT helps with localizing the right answer, although the answers from both the models (base model and HINTed model) are incorrect.

\begin{figure*}[t]
 \centering
 \begin{subfigure}{.5\textwidth}
  \centering
  \includegraphics[scale=0.25]{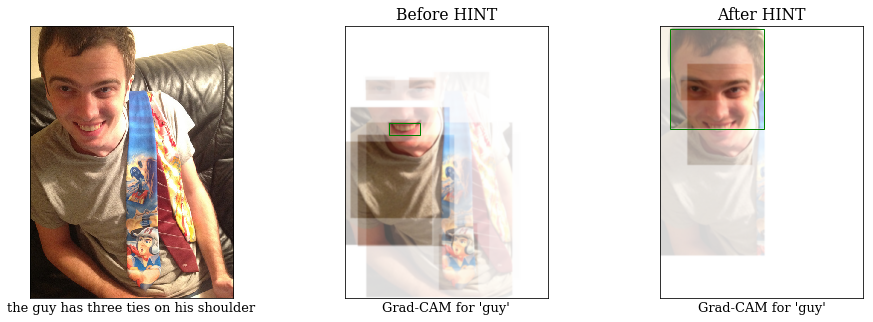}
  \vspace{-10pt}
  \caption{}
\end{subfigure}%
\begin{subfigure}{.5\textwidth}
  \centering
  \includegraphics[scale=0.25]{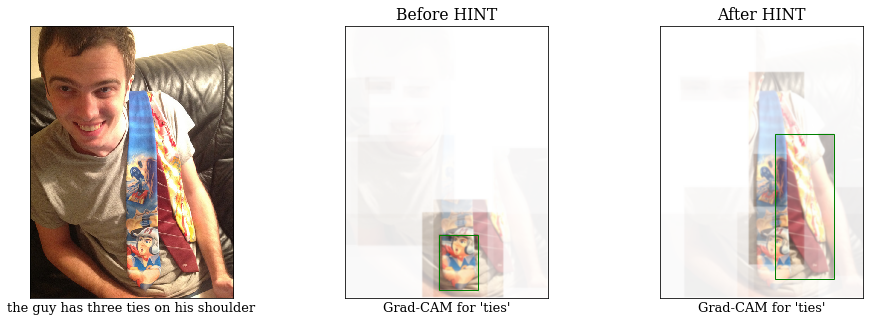}
  \vspace{-10pt}
  \caption{}
\end{subfigure}
\begin{subfigure}{.5\textwidth}
  \centering
  \includegraphics[scale=0.25]{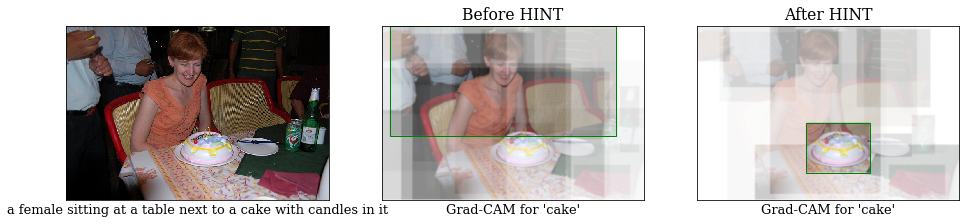}
  \vspace{-10pt}
  \caption{}
\end{subfigure}%
\begin{subfigure}{.5\textwidth}
  \centering
  \includegraphics[scale=0.25]{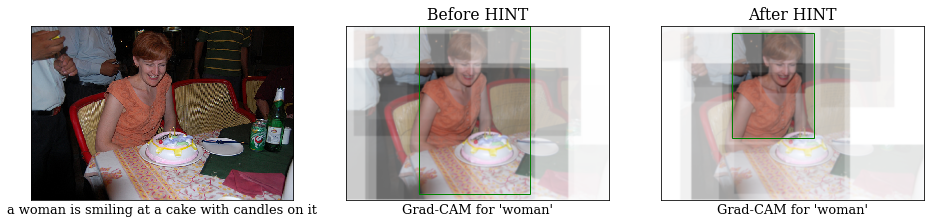}
  \vspace{-10pt}
  \caption{}
\end{subfigure}
\begin{subfigure}{.5\textwidth}
  \centering
  \includegraphics[scale=0.25]{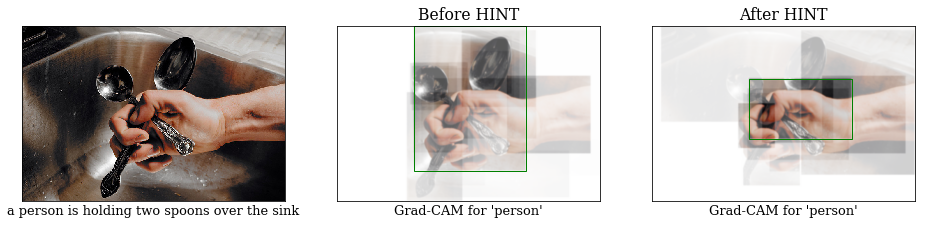}
  \vspace{-10pt}
  \caption{}
\end{subfigure}%
\begin{subfigure}{.5\textwidth}
  \centering
  \includegraphics[scale=0.25]{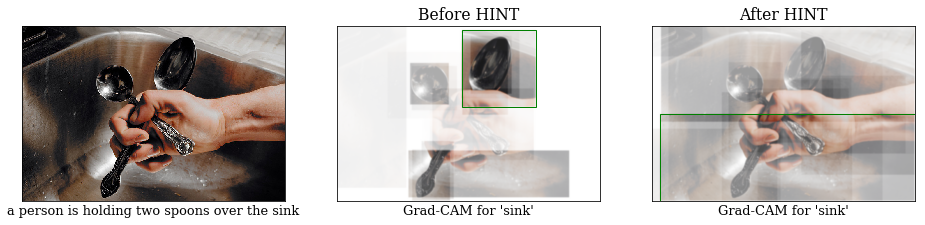}
  \vspace{-10pt}
  \caption{}
\end{subfigure}
\begin{subfigure}{.5\textwidth}
  \centering
  \includegraphics[scale=0.25]{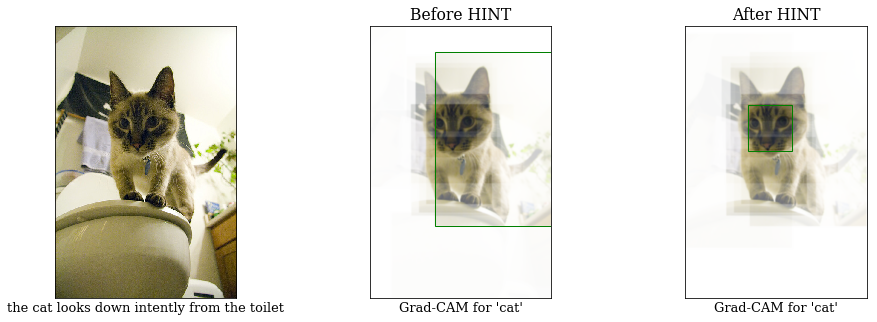}
  \vspace{-10pt}
  \caption{}
\end{subfigure}%
\begin{subfigure}{.5\textwidth}
  \centering
  \includegraphics[scale=0.25]{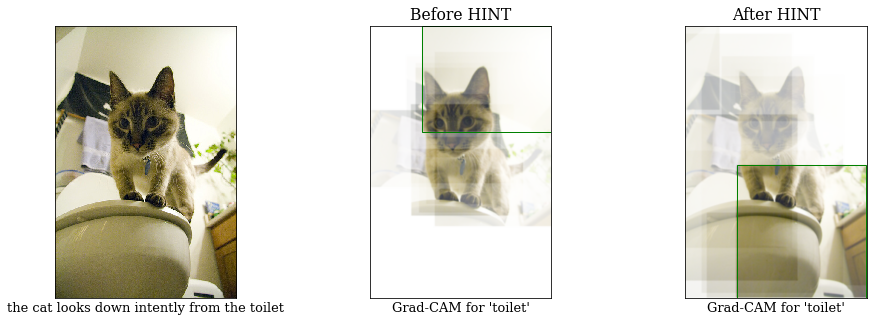}
  \vspace{-10pt}
  \caption{}
\end{subfigure}
\begin{subfigure}{.5\textwidth}
  \centering
  \includegraphics[scale=0.25]{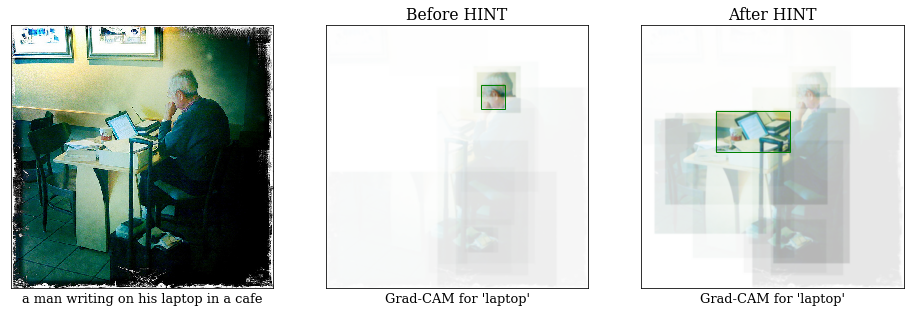}
  \vspace{-10pt}
  \caption{}
\end{subfigure}%
\begin{subfigure}{.5\textwidth}
  \centering
  \includegraphics[scale=0.25]{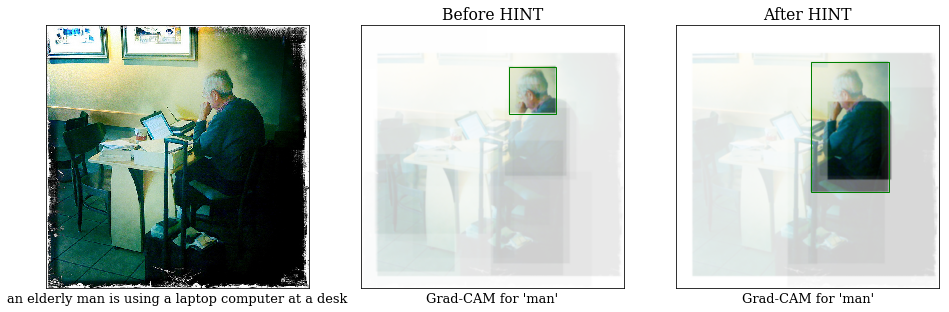}
  \vspace{-10pt}
  \caption{}
\end{subfigure}
\begin{subfigure}{.5\textwidth}
  \centering
  \includegraphics[scale=0.25]{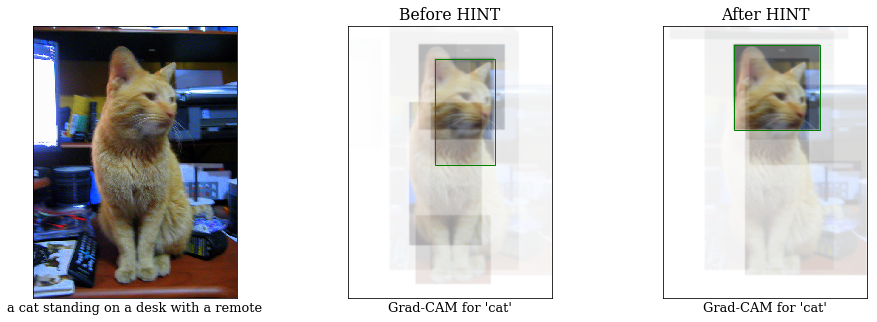}
  \caption{}
\end{subfigure}%
\begin{subfigure}{.5\textwidth}
  \centering
  \includegraphics[scale=0.25]{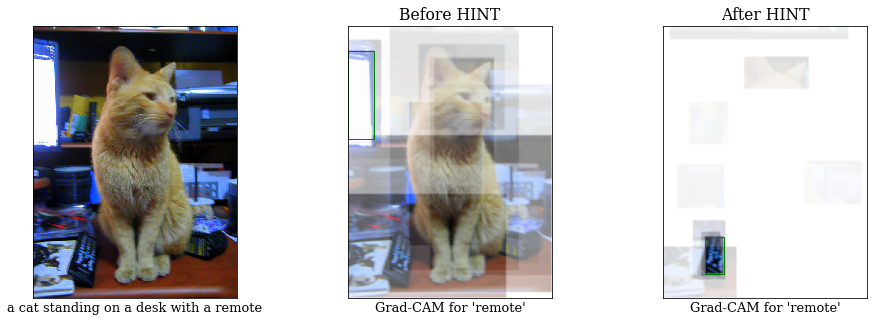}
  \vspace{-10pt}
  \caption{}
\end{subfigure}
\begin{subfigure}{.5\textwidth}
  \centering
  \includegraphics[scale=0.25]{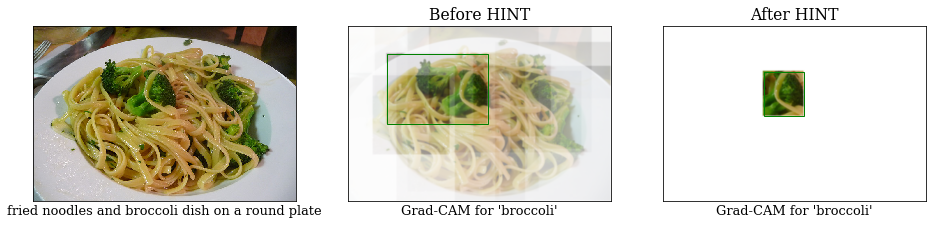}
  \vspace{-10pt}
  \caption{}
\end{subfigure}%
\begin{subfigure}{.5\textwidth}
  \centering
  \includegraphics[scale=0.25]{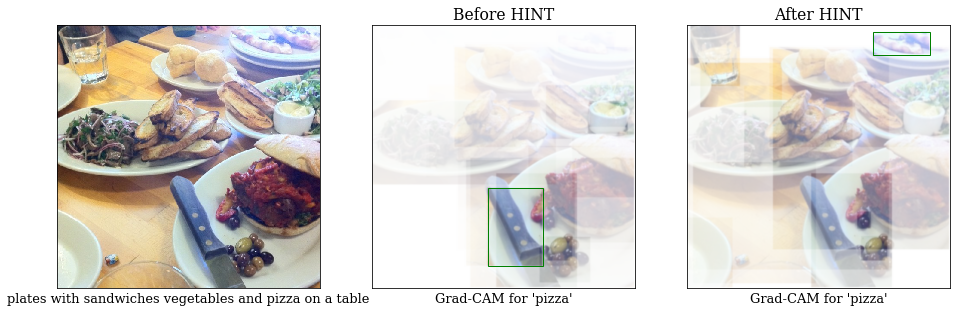}
  \vspace{-10pt}
  \caption{}
\end{subfigure}
\vspace{-10pt}
\caption{Qualitative comparison of Top-down Bottom-up captioning model before and after applying HINT. The left column shows the input image along with the ground-truth caption from the COCO robust split. In the middle column, for the base model we show the explanation visualization for the visual word mentioned below. Similarly we show the explanations for the HINTed models in the third column. We see that the HINTed model looks at more appropriate regions. For example in (k) and (l) note how the HINTed model correctly localizes the visual words `cat' and `remote' accurately when generating the corresponding visual words, but the base model fails to do so. %
}
\vspace{-15pt}\label{fig:qual_captioning}
\end{figure*}

In Fig. \ref{fig:qual_captioning} we show qualitative examples showing the effect of applying HINT on the Top-down Bottom-up \cite{anderson_2017} captioning model trained on the Robust split of the COCO dataset. 
The left column shows the input image along with the ground-truth caption from the COCO robust split. In the middle column, for the base model we show the explanation visualization for the visual word mentioned below. 
Similarly we show the explanations for the HINTed models in the third column. We see that the HINTed model looks at more appropriate regions when generating the mentioned visual word (below the visualization). 

For example, for the input image in Fig. \ref{fig:qual_captioning} (a) and (b), the base model only places a little importance on the face while generating the word `guy', whereas the HINTed model correctly looks at the face of the person. Similarly when generating `ties' the HINTed model looks at the whole tie region compared to the base model. 
Similarly for the images in (e) and (f), the HINTed model looks more correctly at the spoons for the visual word `spoon' and sink, for the word `sink'.

\end{appendices}

{\small
\bibliographystyle{ieee_fullname}
\bibliography{egbib}
}

\end{document}